\documentclass[conference]{IEEEtran}
\IEEEoverridecommandlockouts
\usepackage{amsmath,amsfonts}
\usepackage{amssymb}
\usepackage[noend]{algpseudocode}
\usepackage{algorithm}
\usepackage{array}
\usepackage{textcomp}
\usepackage{tabularx,booktabs}
\usepackage{stfloats}
\usepackage{url}
\usepackage{multirow}
\usepackage{verbatim}
\usepackage{graphicx}
\usepackage{gensymb}
\usepackage{cite}
\usepackage{svg}
\usepackage{subfigure}
\usepackage{xcolor}

\begin{document}
\renewcommand{\baselinestretch}{1}
\title{ Vision Transformer Based \\User Equipment Positioning}

\author{\IEEEauthorblockN{Parshwa Shah\IEEEauthorrefmark{1},
Dhaval K. Patel\IEEEauthorrefmark{2}, Brijesh Soni\IEEEauthorrefmark{3}, Miguel López-Benítez\IEEEauthorrefmark{4}, Siddhartan Govindasamy\IEEEauthorrefmark{5}}
\IEEEauthorblockA{\IEEEauthorrefmark{1} College of Computing and Informatics, University of North Carolina, Charlotte, NC, USA \\ \IEEEauthorrefmark{2}School of Engineering And Applied Sciences, Ahmedabad, Gujarat, India \\ \IEEEauthorrefmark{3}Department of Computer Science and Engineering, The Ohio State University, Columbus, OH, USA \\ \IEEEauthorrefmark{4}School of Computer Science and Informatics, University of Liverpool, Liverpool L69 3DR, United Kingdom\\ \IEEEauthorrefmark{4}ARIES Research Centre, Antonio de Nebrija University, 28015 Madrid, Spain\\ \IEEEauthorrefmark{5}Department of Engineering, Boston College, Chestnut Hill, MA, USA\\}
\IEEEauthorblockA{Email: \IEEEauthorrefmark{1}pshah77@uncc.edu, \IEEEauthorrefmark{2}dhaval.patel@ahduni.edu.in, \IEEEauthorrefmark{3}soni.152@osu.edu, \\ \IEEEauthorrefmark{4}M.Lopez-Benitez@liverpool.ac.uk, \IEEEauthorrefmark{5}siddhartan.govindasamy@bc.edu}
}

\maketitle
\thispagestyle{empty}
\pagestyle{empty}

\begin{abstract}
Recently, Deep Learning (DL) techniques have been used for User Equipment (UE) positioning. However, the key shortcomings of such models is that: i) they weigh the same attention to the entire input; ii) they are not well suited for the non-sequential data e.g., when only instantaneous Channel State Information (CSI) is available. In this context, we propose an attention-based Vision Transformer (ViT) architecture that focuses on the Angle Delay Profile (ADP) from CSI matrix. Our approach, validated on the `DeepMIMO' and `ViWi' ray-tracing datasets, achieves an Root Mean Squared Error (RMSE) of 0.55m indoors, 13.59m outdoors in DeepMIMO, and 3.45m in ViWi's outdoor blockage scenario. The proposed scheme outperforms state-of-the-art schemes by $\sim$ 38\%. It also performs substantially better than other approaches that we have considered in terms of the distribution of error distance.
\end{abstract}

\begin{IEEEkeywords}
Positioning, Localization, Vision Transformer, Attention Mechanism, 5G/6G.
\end{IEEEkeywords}

\section{Introduction}
Positioning is a very \textcolor{black}{demanding use-cases of mobile systems \cite{junglas_location-based_2008}, especially for 5G and beyond}. 
\textcolor{black}{The demand for accurate positioning has grown due to the rise in popularity of IoT devices and industrial automation.} 5G New Radio (NR) 3GPP Rel-16 \cite{3rd_generation_partnership_project_3gpp_5g_nodate} \textcolor{black}{ introduced positioning/localization based features for upcoming wireless network models and has been evolving since then}.
The goal is to enhance network-based positioning, locating User Equipment (UE) via wireless signals for increased accuracy compared to global positioning system alone.

5G localization's accuracy and versatility find applications in various fields, from optimizing traffic and indoor navigation to vehicle tracking. Existing methods are typically algorithmic or data-driven, with algorithm-driven approaches often relying on Time of Arrival (ToA) and Angle of Arrival (AoA) from multiple Base Stations (BS) for triangulation
\cite{jia_motion_2018, papp_tdoa_2022, kakkavas_multi-array_2018}. There are also other methods which work on the basis of the channel matrix such as \cite{li20245g}.
\textcolor{black}{Such methods are limited when there is no Line of Sight (LoS) path available, making accurate position prediction challenging.}

\textcolor{black}{Due to the rapid advancement of learning-based signal processing techniques and the excellent learning ability of data-driven approaches, some works have utilized machine learning/deep learning (ML/DL) techniques for positioning and localization \cite{recent_survey_on_positioning, 10614100}.}
For instance, the work in \cite{vieira_deep_2017} investigated how Convolutional Neural Network (CNN) models can improve accuracy of network-based positioning algorithms. It shows that the Deep-CNN (DCNN) model improves localization accuracy conditioned on the availability of sufficient data points. \textcolor{black}{Another similar work is \cite{sun_fingerprint-based_2019}, where a DCNN approach is considered. This work introduced \textcolor{black}{special} pre-processing as well as multiple deep layers with varying kernel sizes.} 
The authors in \cite{hejazi_dyloc_2021} present a recurrent neural network-based predictive model for LoS and Non-Line of Sight (NLoS) scenarios. Similarly, Received Signal Strength Indicator (RSSI) based UE positioning using Long Short Term Memory (LSTM) architecture is proposed in \cite{miguel_fingerprint_work_2022}. Similarly, the authors in \cite{10466640} showcase work utilizing simple PCA and Bi-LSTM / GRU models for UE positioning. 
However, these works have considered multiple frame inputs where the knowledge of the previous location/history is available. 
\textcolor{black}{In scenarios like, device-to-device and sidelink based system, where the temporal data is not available, such approaches may not render high prediction accuracy.} \textcolor{black}{Additionally, the above architectures extract the spatial/temporal features and weighs the same attention to the entire input data.}

\textcolor{black}{Recently, the attention mechanism has received significant interest from the research community, wherein the key idea is to concentrate only on the relevant information in the data.}
{ \textbf{This is a promising approach, especially for localization and/or positioning problems.} 
There are very few works in the literature that have explored the attention mechanism for localization. For instance, the work in \cite{ruan_hi-loc_2022} utilizes CNN-LSTM for the attention mechanism in spatiotemporal data.} \textcolor{black}{However, it uses sequence-to-sequence distinctive attention mechanism which converts time-series input to more interpretable encoded sequences. Such models are not well suited for the non-sequential data; for instance when the UE is static/quasi-static and/or in the blockage scenario during which only instantaneous Channel State Information (CSI) is available.}

\textcolor{black}{There have been recent advances in the attention mechanism as proven by the transformer models\cite{vaswani2017attention}. Moreover, it is observed that Vision Transformer (ViT)\cite{dosovitskiy_image_2021} is excellent at retrieving spatial features from images. This suggests the idea that if we can map CSI to new embeddings in which spatial features can help predict UE location then we can utilize ViT to significantly improve prediction accuracy.
In this context, we compute the Angle Delay Profile (ADP) which approximates power value of signal against AoA and ToA. 
}\textcolor{black}{Motivated by this idea, we utilize multi-headed self-attention mechanism of ViT for the UE localization.} To the best of the authors' knowledge, UE positioning using a transformer-based attention mechanism, although a promising approach, has not yet been reported in the literature.

\begin{figure*}[t!]
    \centering
    \includegraphics[scale=0.20]{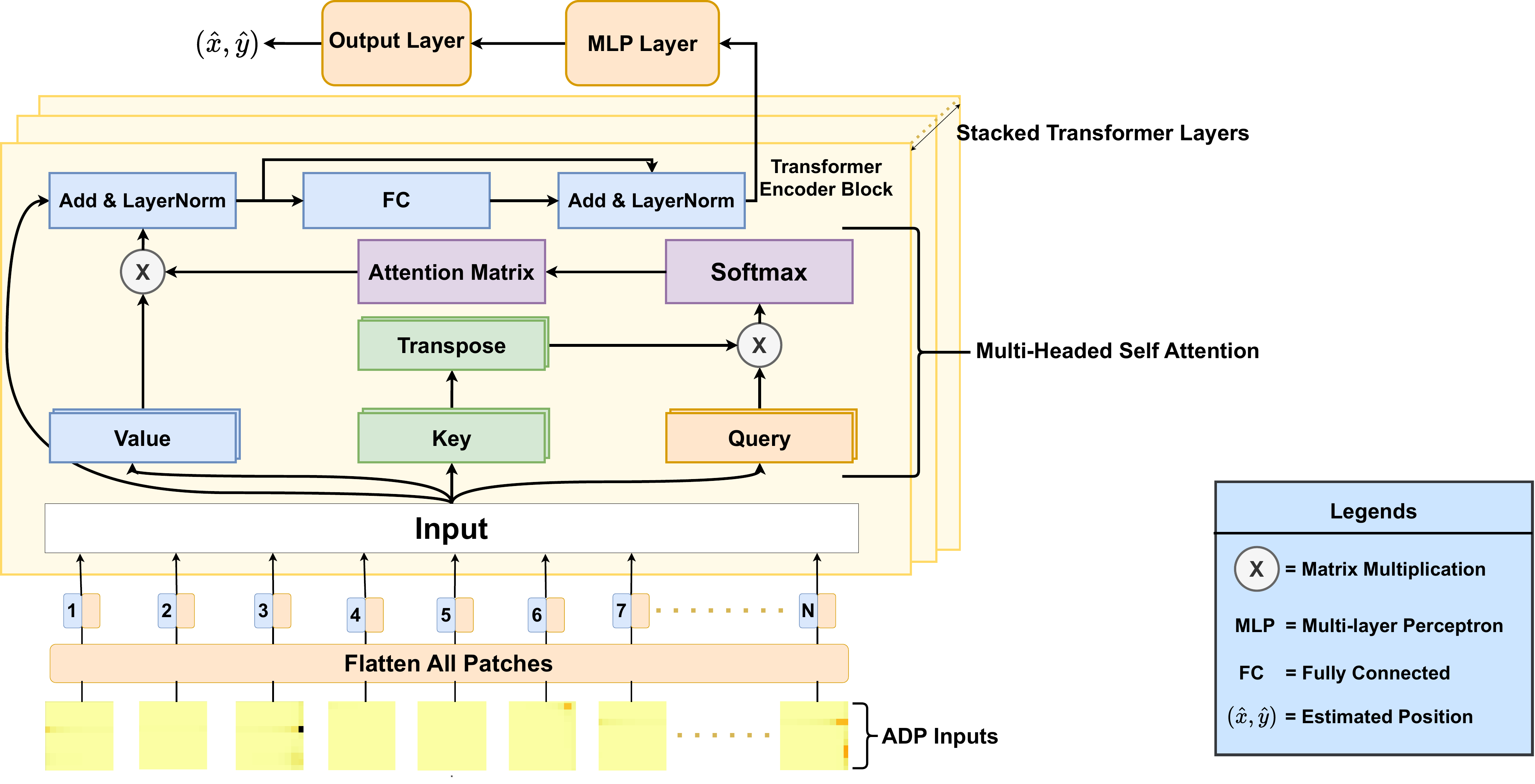}
    \caption{Vision Transformer architecture to process ADP inputs}
    \label{fig:fig5}
\end{figure*}


The key contribution of this work is a transformer\footnote{From now on, we interchangeably use ViT with transformer} model-based novel predictive positioning scheme. Transformer models use positional embeddings, which allow them to learn not only the features in an ADP image but also their spatial locations. This 
allows the model to learn the angle and delay relation in the image.
To validate the robustness of the proposed approach, the model is trained and tested on ray tracing-based DeepMIMO\cite{alkhateeb_deepmimo_2019}, and ViWi\cite{alrabeiah_viwi_2020} datasets. \textcolor{black}{ViWi has a blockage scenario that contains locations where UE has no LoS path with its BS. This type of condition is very likely to happen in urban scenarios. In the DeepMIMO outdoor scenario proposed model achieves \textbf{13.59m} Root Mean Squared Error (RMSE) and for the indoor scenario, the model performs with RMSE of \textbf{0.55m.} In the ViWi scenario, the model has an RMSE of \textbf{3.45m}.} \textcolor{black}{We also compare the Cumulative Distribution Function (CDF) of the prediction error using our approach with other methodologies and find out that it substantially out-performs existing methodologies.}

\textcolor{black}{The rest of this paper is organized as follows. First, Section II describes the system model and preliminaries of the ViT in the context of this work. The proposed scheme, dataset construction, and methodology are comprehensively discussed in Section III. Section IV describes the experimental results. Finally, Section V draws the conclusions from this work.}


\begin{figure*}[t!]
    \centering
    \subfigure[Channel Response Matrix (\textbf{H}) From I (near) to IV (far) shows the blur in overall matrix]{\includegraphics[height=6cm]{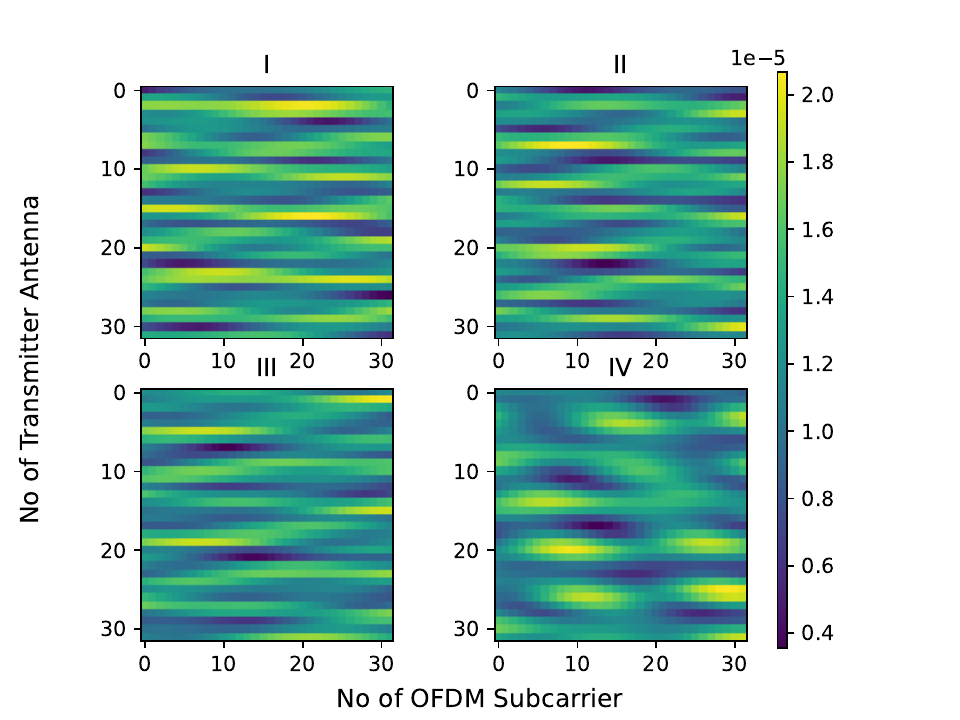}}
    \subfigure[Enhanced Fingerprint used to train model named ADP removes the blur and creates ADP]{\includegraphics[height=6cm]{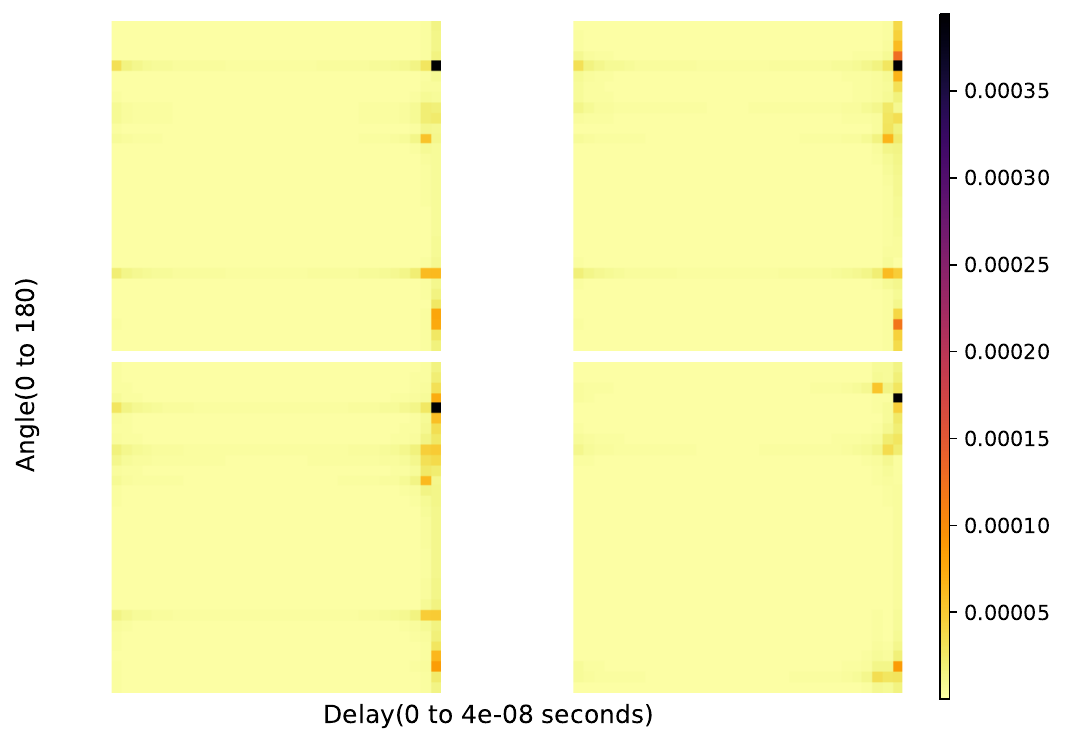}}
    \caption{Visualization of \textbf{H} and ADP}
    \label{fig:Hmatrix}
\end{figure*}

\section{System Model and Preliminaries}

\subsection{System Model}

We assume a UE with a single antenna, served by a single BS\footnote{For indoor scenarios, we refer to the access point as BS.}. Similar to \cite{hejazi_dyloc_2021} and \cite{ali_millimeter_2018}, we assume that BS has a Uniform Linear Array (ULA) antenna and half wavelength spacing between antennas. The BS has $N_t$ antennas which employ Orthogonal Frequency Division Multiplexing (OFDM) with $N_c$ sub-carriers. Furthermore, we assume the channel model with $C$ clusters of users. All clusters have $R_c$ paths. $n_m^{(k)}$ and $\alpha_m^{(k)}$ denotes the delay and a complex gain of $k^{th}$ cluster and $m_{th}$ path of that cluster, respectively.  The channel coefficient of $l^{th}$ sub-carrier is $ h[l], $ which can be written as \cite{alkhateeb_frequency_2016}:
\begin{equation}
    h[l]=\sum_{k=1}^C \sum_{m=1}^{R_C} \alpha_m^{(k)} e\left(\theta_m^{(k)}\right) e^{-j 2 \pi \frac{l n_m^{(k)}}{N_c}},
    \label{eq:1}
\end{equation}

where $\mathbf{e}(\theta)$ is the response vector of the ULA antenna employed by the BS in the assumed scenario
\begin{equation}
    \mathbf{e}(\theta)=\left[1, e^{-j 2 \pi \frac{d \cos (\theta)}{\lambda}}, \ldots, e^{-j 2 \pi \frac{\left(N_t-1\right) d \cos (\theta)}{\lambda}}\right]^T,
    \label{eq:2}
\end{equation}

Here, $\theta$ is the angle of arrival and $\lambda$ denotes the wavelength of the carrier signal. Combining all sub-carrier responses in one matrix gives a CSI matrix ({$\textbf{H} \in \mathbb{C}^{N_t \times N_c}$}), which can be expressed as:
\begin{equation}
    \textbf{H}= \left[h[1], \ldots, h\left[N_c\right]\right].
    \label{eq:3}
\end{equation}

\textcolor{black}{This encapsulates the whole channel modeling part of the system model. Now the problem statement translates to creating a model which is capable of regressing UE location from $\mathbf{H}$ (\ref{eq:3}). We further plan to use ViT for this problem. Preliminaries of the same are addressed in next sub-section.}

\subsection{Preliminaries of Vision Transformer}
The ViT architecture \cite{dosovitskiy_image_2021} is as shown in Fig. \ref{fig:fig5}, which briefly describes the attention mechanism. We first convert the CSI matrix to the ADP matrix, the details of which are provided in the upcoming sections.

Here we process the ADP as an image through the ViT model. The model first divides the ADP matrix into equal patches, shown as ADP inputs in Fig. \ref{fig:fig5}. A patch embedding layer is applied in the transformer to extract location-specific data from the ADP matrix. Then these patches are flattened and embedded with position tokens. The patch size is fixed at $6 \times 6$, which means the ADP matrix is divided into `\textbf{N}' number of patches of the same size, which are shown in blue-orange boxes in Fig. \ref{fig:fig5}.
These \textcolor{black}{embedded} tokens are then converted to multi-headed key (shown in green), value (blue), and query (orange) matrices. We transpose the key matrix and multiply it with the query matrix, potentially producing high values. To suppress them and to convert the matrix to probabilities, a softmax layer is used. The matrix received at the end is known as an attention matrix or attention mask (purple). This matrix indicates which parts of the input image the model should concentrate on. Then, the attention matrix is multiplied with a values matrix to create a residual connection with the input and is then passed to the normalization layer (This is referred to in the figure as `Add \& LayerNorm'.).
The normalized output is then sent to the fully connected (FC) layer, again passed to Add \& LayerNorm (This section is referred as transformer encoder block.). It is stacked to extract the maximum information from the input as shown by stacking yellow boxes in the background. The final output of the transformer encoder blocks is passed through a multi-layer perceptron (MLP) model to generate the position prediction. 

\textcolor{black}{The ability to focus on each patch differently allows the ViT model to get the details about prominent ToA and AoA pairs related to the original signal. This information is finally embedded in the resulting vector, which is then used to regress the UE's predicted location ($\hat{x}, \hat{y}$).}
\textcolor{black}{ViT-based prediction helps in generating an accurate mapping of the ADP matrix with the UE location, which is described in the next section.}




\section{Proposed Approach and Methodology}
This section discusses data pre-processing, the generation of data for model-based learning, and proposed approach.

\subsection{Data Preprocessing}
We pre-process \textbf{H} using (\ref{eq:3}), which results in an ADP \cite{sun_fingerprint-based_2019}.
\textcolor{black}{This ADP makes it possible for the transformer model to easily understand \textbf{H} and predict precise UE's location from it.}
ADP is achieved with two Discrete Fourier Transform (DFT) matrices. The first one is defined as $\textbf{V} \in \mathbb{C}^{N_t \times N_t} $:
\begin{equation}
    [\textbf{V}]_{z, q} \triangleq \frac{1}{\sqrt{N_t}} e^{-j 2 \pi \frac{\left(z\left(q-\frac{N_t}{2}\right)\right)}{N_t}} ,
    \label{eq:4}
\end{equation}
and another is defined as $\textbf{F} \in \mathbb{C}^{N_c \times N_c} $
\begin{equation}
    [\textbf{F}]_{z, q} \triangleq \frac{1}{\sqrt{N_c}} e^{-j 2 \pi \frac{z q} {N_c}} ,
    \label{eq:5}
\end{equation}
\textcolor{black}{where} $[X]_{z,q}$ means item on the $z, q$-th entry of the matrix X. Finally, we multiply the DFT matrices with $\textbf{H}$ to obtain the ADP matrix
as below,
\begin{equation}
    \textbf{A} = \mid \textbf{V}^H \textbf{H F} \mid
    \label{eq:6}
\end{equation}
Here, $\textbf{A}$ is the ADP matrix, which is used to get a more detailed understanding of \textbf{H} (CSI matrix).

As shown in Fig. \ref{fig:Hmatrix}, \textbf{H} matrix has a set of values that do not appear very structured, and from which it is difficult to infer location information. \textcolor{black}{Subplot I in Fig. \ref{fig:Hmatrix}(a) shows CSI of the UE in subplot-I when it is closest to BS and in subplot-IV when it is farthest. As the UE moves away, \textbf{H} becomes more blurred, making it difficult for the DL model to understand.} Fig. \ref{fig:Hmatrix}(b) shows an angle v/s delay plot showing the power of the path containing delay and angle information, respectively. It can be seen that regardless of the distance, the ADP matrices corresponding to the subplots in Fig. \ref{fig:Hmatrix}(a) are much clearer. \textcolor{black}{This pre-processing task makes the model learning faster and more robust.\cite{sun_fingerprint-based_2019}} The use of DFT converts \textbf{H} from being in the dimension of transmitter and sub-carriers to the domain of angle and delay.
This helps the model to easily learn location identification from ADP compared to the same with \textbf{H}.
Nonetheless, the performance improvement is still due to the attention mechanism of the transformer, as will be revealed in the results section.

\subsection{Dataset generation}

Acquiring real-time data for the 3GPP specification of 5G NR specifically for positioning problems is difficult. Collecting real-time data from commercial off-the-shelf transceivers is extremely challenging as this data is typically inaccessible.  
To overcome these limitations, we explore and use the ray-tracing-based simulations from DeepMIMO \cite{alkhateeb_deepmimo_2019} and ViWi \cite{alrabeiah_viwi_2020}, which capture both indoor and outdoor settings.

DeepMIMO\cite{alkhateeb_deepmimo_2019} can generate ray-tracing data with CSI for various locations. ViWi \cite{alrabeiah_viwi_2020} uses the same parameters as DeepMIMO based on 3GPP 5G-NR Rel 16 technical specifications \cite{3rd_generation_partnership_project_3gpp_5g_nodate}. It has a Clustered Delay Line (CDL) channel model for the simulation of outdoor scenarios. 
The dataset generation parameters are as shown in Table \ref{tab:table1}. For robust comparison, we generate three scenarios: i) DeepMIMO indoor, ii) DeepMIMO outdoor, and iii) ViWi outdoor blockage datasets. These include comprehensive data with close-to-realistic simulation methodologies and these approaches have been used in numerous other works utilizing ML not just in positioning contexts (\cite{vuckovic_map-csi_2021, sharma_fedbeam_2023, hejazi_dyloc_2021}).

\begin{table}[t!]
\centering
\caption{Dataset Generation parameters\label{tab:table1}}
\renewcommand{\arraystretch}{1}
\begin{tabular}{|c|c|}
    \hline
    \textbf{Parameter} & \textbf{Value (Outdoor/Indoor)} \\
    \hline
    Carrier Frequency & 3.5 GHz / 60 GHz \\ \hline
    No. of Resource Block & 64 / 32 \\ \hline
    BS Antenna Size (ULA) & 1 x 64 / 32 distributed antennas \\ \hline
    UE Antenna Size & 1 x 1 \\ \hline
    BS Orientation & 0\degree , 0\degree \\ \hline
    UE Azimuth \& Elevation Range & [0\degree, 360\degree] \& [0\degree, 90\degree] \\ \hline
    No. of paths ($R_c$) & 25 \\ \hline
\end{tabular}
\end{table}
This simulation is used to generate 49,500 data points in Outdoor Scenario; 28,393 data points in Indoor Scenarios; and 5,000 ViWi data points which include \textit{blockages}. These points are generated assuming that a UE is at a random location in pre-defined scenarios. \textcolor{black}{The number of transmitter antennas and the sub-carriers for all generated scenarios are 64, 32, and 60, respectively.} These numbers match with previous works to which the proposed model is compared. 
\textcolor{black}{From the generated data, 64\% is used for training, 16\% for validation, and 20\%  for testing.}
The ViT model proposed in this paper uses the mean squared error as
loss function and uses RMSE to quantify the predictions. The label set for the proposed model is UE position in
(x, y) coordinates.

\subsection{Proposed Approach}
Initially, the simulation environment is set up by defining parameters such as 3D map, BS \& UE settings. These settings generate a CSI matrix which is then converted to the ADP. It is then passed on to the model to learn location features from it.
Algorithm \ref{algo:generation} defines how the data were collected in all three scenarios. The algorithm describes two procedures, namely: i) channel generation, and ii) training. In channel generation, a 3D model of the scenario, BS, and UE settings are imported into a ray tracing environment. This generates a channel matrix for all different UE positions using CDL modeling, which collects the delay and noise of different rays between UE and BS. The channel matrix is then multiplied with $\textbf{V}$ \& $\textbf{F}$ as in (\ref{eq:4}), (\ref{eq:5}) which creates ADP matrix ($\mathbf{A}$).
The ADP matrix contains power profile for angle and delay pairs.
Then a ViT model is created as mentioned in Section. II.B.
\textcolor{black}{The model utilizes the ADP matrix as an image input, employing an attention mask from the ViT. During processing, it seeks high-power patches to ascertain accurate location information.}
\textcolor{black}{The true potential of this model lies in its ability local intese signal directions and follow corresponding angle and delay from $\mathbf{A}$. This is achieved using the attention mechanism as explained earlier.}

\begin{algorithm}[t]
\caption{Data Generation, Simulation and Training Phase}
\label{algo:generation}
\begin{algorithmic}[1]

\Procedure{\textcolor{black}{Channel Generation}}{}
    \State Import 3D Model, Select locations of UE and BS 
    \State Run ray tracing for the settings specified
    \State Divide UE in clusters of same delay and different AoA
    \For{$l \in$ sub-carriers}
        \For{$k \in$ clusters}
            \For{$m \in$ rays}
                \State Collect $\alpha_m^{(k)}$ and $n_m^{(k)}$ for Eq. \ref{eq:1}
            \EndFor
            \State Calculate Eq. \ref{eq:1}
        \EndFor
        \State Calculate $h[l]$ using Eqs. \ref{eq:1} \& \ref{eq:2}
    \EndFor
    \State Concatenate to generate final \textbf{H} as in Eq. \ref{eq:3}
\EndProcedure

\Procedure{Training}{channels, locations}
\State $\textbf{V}=$ Eq. \ref{eq:4} where $N_t=$ No. of Transmitters
\State $\textbf{F}=$ Eq. \ref{eq:5} where $N_c=$ No. of Sub-Carriers
\State model = Proposed DL model
\For{$_ \in$ epochs}
\For{$(\textbf{H}, p) \in$ (channels, locations)}
    \State $ \textbf{H}_p = \textbf{V}^H \textbf{H F}$ as in Eq. \ref{eq:6}
    \State $\textbf{A}=| \textbf{H}_p |$
    \State Train model to learn $\textbf{A}$, validate and test
\EndFor
\EndFor
\EndProcedure
\end{algorithmic}
\end{algorithm}

\section{Experimental Results}

The experiments were carried out using Tensorflow\cite{tensorflow2015-whitepaper} GPU API on a dedicated Param Shavak high-performance super computing system. Using the proposed approach, we implemented the position prediction model and tested it on DeepMIMO \cite{alkhateeb_deepmimo_2019} and ViWi \cite{alrabeiah_viwi_2020} datasets.
Model hyperparameters are described in Table \ref{tab:params}.

\begin{table}[b!]
\centering
\caption{Proposed Model hyperparameters}
\label{tab:params}
\begin{tabular}{|c|c|}
    \hline
    \textbf{Parameter} & \textbf{Value} \\
    \hline
    Epochs; Learning Rate; Weight Decay & 50; 0.001; 0.0001 \\ \hline
    Loss Function & Mean Squared Error (MSE) \\ \hline
    ADP size (Outdoor, Indoor, ViWi) & (64x64), (32x32), (60x60) \\\hline
    Patch Size & 6 x 6 \\ \hline
    Number Of Heads; Transformer Layers & 4; 8 \\ \hline
    MLP Sizes & 128, 64 \\
    \hline
\end{tabular}
\end{table}

\textcolor{black}{We compare the proposed approach with the state-of-the-art schemes, \textbf{i) DCNN \cite{sun_fingerprint-based_2019}:} In this scheme authors have used DCNN model with varying kernel sizes along with ADP based pre-processing over CSI Matrix; \textbf{ii) FedBeam \cite{sharma_fedbeam_2023}:} In this work, authors use multiple ray characteristics like AoA, ToA, RSSI, etc. in a federated learning network to protect users' privacy while sharing location information to create new beam; \textbf{iii) MAP-CSI \cite{vuckovic_map-csi_2021}:} In this approach, authors have utilized environment map along with AoA and ToA to further assist the model to predict more accurate location. \textbf{We would like to highlight that all the above schemes require additional data dedicated for positioning method. On the contrary, our proposed method utilizes ADP (which only requires CSI) and attention based ViT to predict UE's location.}} 

\begin{figure}[t!]
    \centering
    \includegraphics[height=7cm,width=\linewidth]{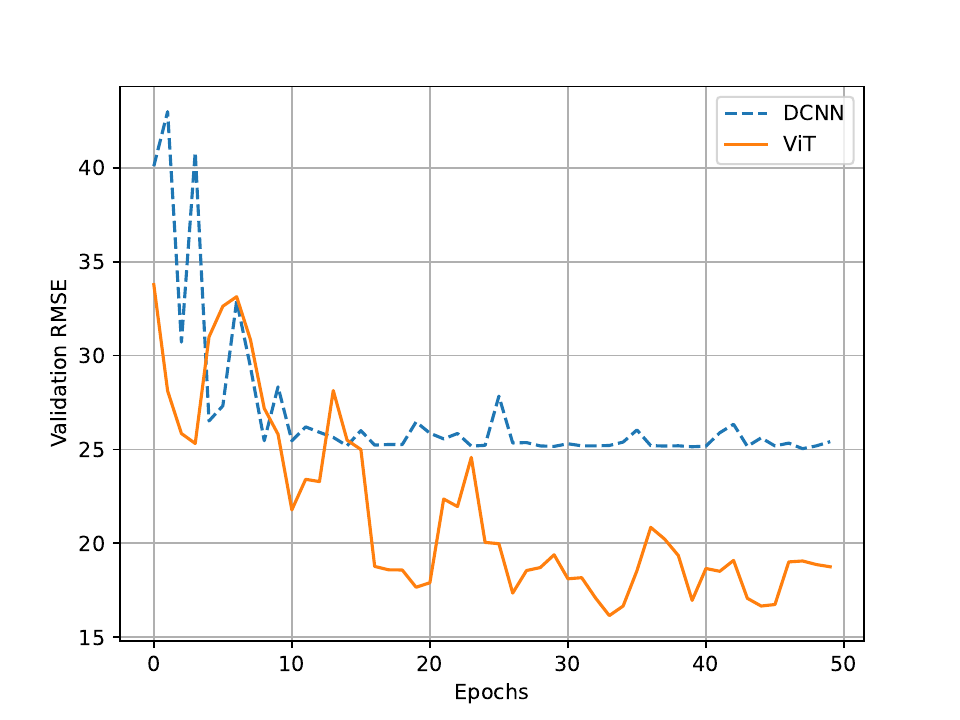}
    \caption{DCNN and ViT Validation RMSE Comparison}
    \label{fig:fig6}
\end{figure}


In Fig. \ref{fig:fig6}, ViT is compared with DCNN to quantify the validation RMSE over the epochs. We can notice that the performance of ViT is better than the DCNN model \cite{sun_fingerprint-based_2019}. This is because the transformer assigns higher attention weights to the most relevant parts of the input, enabling it to extract features more effectively, which in turn allows the ViT architecture to learn faster even with the same amount of data.

\begin{figure}[t!]
    \centering
    \includegraphics[height=6.5cm,width=\linewidth]{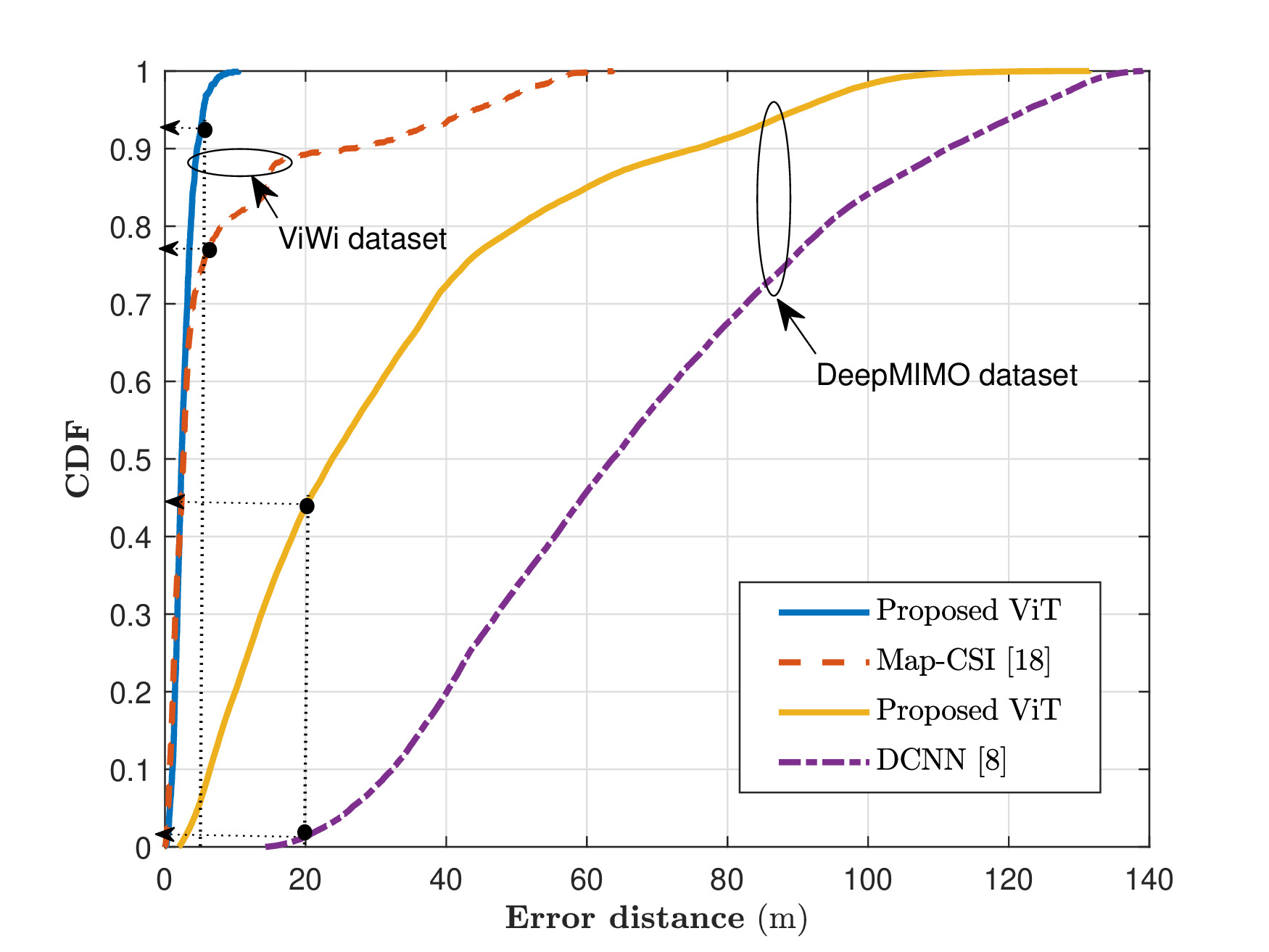}
    \caption{CDF of the error distance}
    \label{fig:learning}
\end{figure}

\textcolor{black}{Moreover, to quantify the distribution of the predictions, we plotted the CDF of the error distances from our ViT approach and compared it with the DCNN \cite{sun_fingerprint-based_2019} and Map-CSI\cite{vuckovic_map-csi_2021} approaches in Fig. \ref{fig:learning}. 
The obtained results indicate that proposed ViT approach performs significantly better than the DCNN and Map-CSI approaches. For instance, for the DeepMIMO Outdoor scenario, the probability that the error is greater than 20m (shown by the dotted arrow) with our approach is approximately $56\%$ compared to the DCNN model where the probability is $\sim98\%$. Similarly, for the 
MAP-CSI model using the ViWi data set, the probability that the error is greater than 5m is $26\%$ as compared to our approach where it is $8\%$.
} \textcolor{black}{This result highlights that the proposed approach performs substantially better than the other schemes in terms of positioning accuracy, even in case of blockage scenario (ViWi).}

    The comparison with these models is shown in Table \ref{tab:comparison}. The selected OFDM parameters (No. of OFDM sub-carriers, Sub-carrier spacing) are the same as used in respective works. For a fair comparison, \textcolor{black}{in outdoor scenarios, we have considered the O1 scenario of DeepMIMO with BS-2 activated, and rows R1 to R1100 in user grid 1, which is also considered in \cite{hejazi_dyloc_2021} to validate DCNN model\cite{sun_fingerprint-based_2019}. Similarly, for indoor simulations, the I3 scenario is considered with user grid 1 being activated completely along with BS-2. Furthermore, the model is validated on DeepMIMO \cite{alkhateeb_deepmimo_2019} where, in outdoor scenarios, the model had positioning error of \textbf{13.59 meter}, which is 45\% better than \cite{sun_fingerprint-based_2019}'s DCNN model, while the indoor scenario has a positioning error of \textbf{0.55 meter}. The model is also applied on ViWi-Dataset from \cite{alrabeiah_viwi_2020}, where also it gave a test error of \textbf{3.45 meter}, which is 31\% better than MAP-CSI proposed in \cite{vuckovic_map-csi_2021}. The proposed model outperforms all the systems made for specific models. Overall, the proposed approach  performs \textbf{$\sim$ 38\%} better as compared to the state-of-the-art schemes.} \textcolor{black}{The 3GPP Rel-17 technical specifications for location services \cite{3gpp_rel_17} mentions different required accuracies for different services. In the outdoor scenarios, the most accurate bracket is 10-50m which is achieved in our simulation.}

\begin{table}[t]
\centering
\caption{Comparison with the State-of-the-art schemes.}
\scalebox{0.9}{
\begin{tabular}{lcccc}
\hline
\textbf{Models} & \textbf{Input Data} & \textbf{Method Used} & \textbf{Scenario} & \textbf{RMSE} \\
\hline
DCNN           & CSI       & DCNN        & DeepMIMO Outdoor & 25m \\
Proposed       & CSI       & Transformer & DeepMIMO Outdoor & \textbf{13.59m} \\
\hline
FedBeam        & CSI       & NN          & DeepMIMO Indoor  & 0.6m \\
Proposed       & CSI       & Transformer & DeepMIMO Indoor  & \textbf{0.55m} \\
\hline
MAP-CSI        & MAP, CSI  & Map-AT      & ViWi Blockage    & 5m \\
Proposed       & CSI       & Transformer & ViWi Blockage    & \textbf{3.45m} \\
\hline
\end{tabular}
}
\label{tab:comparison}
\end{table}
The key reason for the transformer to perform well is that it first breaks the image (ADP) into patches. All these patches are embedded with location details and fed to the transformer. These input patches are then multiplied with the attention matrix using transformers, which allows the model to concentrate on important parts of the ADP matrix. Transformer makes it particularly powerful because all input ADP matrices contain information about power from a certain angle with some delay. Better feature extraction is made possible by the use of transformers, and more precise user position prediction is made possible by the inclusion of location-specific data.

\section{Conclusion}

In this work, we have proposed a transformer model-based novel predictive positioning scheme. A ViT model with a multi-headed self-attention mechanism is considered to find relevant parts of input data to learn from.
We have utilized a pre-processing method to convert the CSI matrix to an ADP matrix, which makes it easier for the model to extract information.
To validate our work, we have used publicly available `DeepMIMO' and `ViWi' datasets with blockage. From DeepMIMO, two scenarios are considered namely indoor (I3) and outdoor (O1). Experimental results have revealed that our proposed model performs 30\% to 40\% better than the state-of-the-art schemes. It also  performs substantially better in terms of the distribution of error distance, even for the blockage scenario. The proposed approach achieves high accuracy and reliability in positioning, using only ADP matrix and without any additional data,
making it a promising solution for various positioning use cases in
beyond 5G/6G networks. 

\bibliographystyle{IEEEtran}
\bibliography{5GLocalization}

\end{document}